\documentclass[a4paper,11pt]{article}

\usepackage{graphicx}  
\usepackage{url}       
\usepackage{times}
\usepackage{natbib}
\usepackage[margin=25mm]{geometry}

\usepackage{caption}

\newtheorem{definition}{Definition}

\title{Using Inverse $\lambda$ and Generalization to Translate English to Formal Languages\footnote{To appear in Ninth International Conference on
 Computational Semantics (IWCS 2011)}}
\date{}

\author{Chitta Baral\\
        Arizona State University\\
        \texttt{chitta@asu.edu}
   \and Juraj Dzifcak\\
        Arizona State University\\
        \texttt{juraj.dzifcak@asu.edu}
   \and Marcos  Alvarez Gonzalez\\
        Arizona State University\\
        \texttt{malvar@asu.edu}
   \and Jiayu Zhou\\
         Arizona State University\\
        \texttt{Jiayu.Zhou@asu.edu}
}

\begin{document}
\maketitle


\begin{abstract}

We present a system to translate natural language sentences to formulas in a formal or a knowledge representation language. Our system uses two
inverse $\lambda$-calculus operators and using them can take as input the semantic representation of some words, phrases and sentences and from that derive the semantic  representation of other words and phrases. Our inverse $\lambda$ operator works on many formal languages including first order logic, database query languages and answer set programming. Our system uses a syntactic combinatorial categorial parser to parse natural language sentences and also to construct the semantic meaning of the sentences as directed by their parsing. The same parser is used for both. In addition to the inverse $\lambda$-calculus operators, our system uses a notion of generalization to learn semantic representation of words from the semantic representation of other words that are of the same category. Together with this, we use an existing statistical learning approach to assign weights to deal with multiple meanings of words. Our system produces improved results on standard corpora on natural language interfaces for robot command and control and database queries.

\end{abstract}


\section{Introduction}

Our long term goal is to develop general methodologies to translate natural language text into a formal knowledge representation (KR) language.  In the absence of a single KR language that is appropriate for expressing all the nuances of a natural language, currently, depending on the need different KR languages are used.  For example, while first-order logic is appropriate for mathematical knowledge, one of its subset Description logic is considered appropriate for expressing ontologies, temporal logics are considered appropriate for expressing goals of agents and robots, and various non-monotonic logics have been proposed to express common-sense knowledge. Thus, one of of our goals in this paper is to develop general methodologies that can be used in translating natural language to a desired KR language.

There have been several learning based approaches, mainly from two groups at MIT and Austin. These include the following works: \cite{Collins:2005}, \cite{Mooney:2006}, \cite{Mooney:2006a}, \cite{Mooney:2007}, \cite{Lu:2008}, \cite{Collins:2007} and \cite{Mooney:2009}. Given a training corpus of natural language sentences coupled with their desired representations, these approaches learn a model capable of translating sentences to a desired meaning representation. For example, in the work by \cite{Collins:2005}, a set of hand crafted rules is used to learn syntactic categories and semantic representations of words based on combinatorial categorial grammar (CCG), as described by \cite{Steedman:Book}, and $\lambda$-calculus formulas, as discussed by \cite{ltfgamut91}. The later work of \cite{Collins:2007}, also uses  {\em hand crafted rules}. The Austin group has several papers over the years. Many of their works including the one by \cite{Mooney:2009} use a word alignment method to learn semantic lexicon and learn rules for composing meaning representation.

Similar to the work by \cite{Mooney:2009}, we use an existing syntactic parser to parse natural language. However we use a CCG parser, as described by \cite{CCG}, to parse sentences, use lambda calculus for meaning representation, use the CCG parsing to compose meaning and have an initial dictionary. Note that unlike the work by \cite{Mooney:2009}, we do not need to learn rules for composing meaning representation. We use a novel method to learn semantic lexicon which is based on two inverse lambda operators that allow us to compute $F$ given $G$ and $H$ such that $F @ G = H$ or $G @ F = H$. Compared to the work by \cite{Collins:2005}, we use the same learning approach but use a completely different approach in lexical generation. Our inverse $\lambda$ operator has been tested to work for many languages including first order logic, database query language, CLANG by \cite{Chen:2003}, answer set programming (ASP) as described by \cite{Baral:Book}, and temporal logic. Thus our approach is not dependent on the language used to represent the semantics, nor limited by a fixed set of rules. Rather, the new $\lambda$-calculus formulas and their semantic models, corresponding to the semantic or meaning representations, are directly obtained from known semantic representations which were provided with the data or learned before. The richness of $\lambda$ calculus allows us to rely only on the syntactic parse itself without the need to have separate rules for composing the semantics. The provided method yields improved experimental results on existing corpora on robot command and control and database queries.


\section{Motivation and Background}

We now illustrate how one can use CCG parsing and $\lambda$-calculus applications to obtain database query representation of sentences. We then motivate and explain the role of our ``inverse $\lambda$'' operator. A syntactic and semantic parse tree for the sentence ``Give me the largest state.'' is given in Table \ref{tab:ex-John2}.

\begin{table*}[htb]
\small{
\begin{center}
\begin{tabular}{c c c c}
Give me & the & largest & state.\\
$S/NP$ & $NP/N$ & $N/N$ & $N$ \\
\cline{3-4}
$S/NP$ & $NP/N$ & $N$ & \\
\cline{2-4}
$S/NP$ & $NP$ & & \\
\cline{1-4}
$S$ & & & \\
\end{tabular}

\begin{tabular}{c c c c}
Give me & the & largest & state.\\
$\lambda x. answer(A,x@A)$ & $\lambda x. x$ & $\lambda x. \lambda y. largest(y, x@y)$ & $\lambda z. state(z)$ \\
\cline{3-4}
$\lambda x. answer(A,x@A)$ & $\lambda x. x$ & $\lambda y. largest(y, state(y))$ & \\
\cline{2-4}
$\lambda x. answer(A,x@A)$ & $\lambda y. largest(y, state(y))$ &  & \\
\cline{1-4}
$answer(A,largest(A, state(A)))$ & & & \\
\end{tabular}

\end{center}
}
\caption{CCG and $\lambda$-calculus derivation for ``Give me the largest state.''}
\label{tab:ex-John2}
\end{table*}

The upper portion of the figure lists the nodes corresponding to the CCG categories which are used to syntactically parse the sentence. These are assigned to each word and then combined using combinatorial rules, as described by \cite{Steedman:Book}, to obtain the categories corresponding to parts of the sentence and finally the complete sentence itself. For example, the category for ``largest'', $N/N$ is combined with the category of ``state.'', $N$, to obtain the category of ``largest state.'', which is $N$. In a similar manner, each word is assigned a semantic meaning in the form of a $\lambda$-calculus formula, as indicated by the lower portion of the figure. The language used to represent the semantics of words and the sentence is the database query language used in the robocup domain. The formulas corresponding to words are combined by applying one to another, as dictated by the syntactic parse tree to obtain the semantic representation of the whole sentence. For example, the semantics of ``the largest state.'', $\lambda y. largest(y, state(y))$ is applied to the semantics of ``Give me'', $\lambda x. answer(A,x@A)$, to obtain the semantics of ``Give me the largest state.'', $answer(A,largest(A, state(A)))$.

The given example illustrates how to obtain the semantics of the sentence given the semantics of words. However, what happens if the semantics of the word ``largest'' is not given? It might be either missing completely, or the current semantics of ``largest'' in the dictionary might simply not be applicable for the sentence ``Give me the largest state.''.

Let us assume that the semantic representation of ``largest'' is not known, while the semantic representation of the rest of the sentence is known. We can then obtain the semantic representation of ``largest'' as follows. Given the formula $answer(A,largest(A, state(A)))$ for the whole sentence ``Give me the largest state.'' and the formula $\lambda x. answer(A,x@A)$ for ``Give me'', we can perform some kind of an {\it inverse application} \footnote{Thus instead of applying $G$ to $F$ to obtain $H$, $G @ F = H$, we try to find an $F$ such that $G@F = H$ given $G$ and $H$.} to obtain the semantics representation of ``the largest state'', $\lambda y. largest(y, state(y))$. Similarly, we can then use the known semantics of ``the'', to obtain the semantic representation of ``largest state.'' as $\lambda y. largest(y, state(y))$. Finally, using the known semantics of state, $\lambda z. state(z)$ we can obtain the the semantics of ``largest'' as $\lambda x. \lambda y. largest(y, x@y)$.

It is important to note that using $@$ we are able to construct relatively complex semantic representations that are properly mapped to the required syntax.

Given a set of training sentences with their desired semantic representations, a syntactic parser, such as the one by \cite{CCG}, and an initial dictionary, we can apply the above idea on each of the sentences to learn the missing semantic representations of words. We can then apply a learning model, such as the one used by \cite{Collins:2005}, on these new semantic representations and assign weights to different semantic representations. These can then be used to parse and represent the semantics of new sentences. This briefly sums up our approach to learn and compute new semantic representations. It is easy to see that this approach can be applied with respect to any language that can be handled by  ``inverse $\lambda$'' operators and is not limited in the set of new representations it provides.

We will consider two domains to evaluate our approach. The fist one is the GEOQUERY domain used by \cite{Zelle:1996}, which uses a Prolog based language to query a database with geographical information about the U.S. It should be noted that this language uses higher-order predicates. An example query is provided in Table \ref{tab:ex-John2}. The second domain is the ROBOCUP domain of \cite{Chen:2003}. This is a multi-agent domain where agents compete against each other in a simulated soccer game. The language CLANG of \cite{Chen:2003} is a formal language used to provide instructions to the agents. An example query with the corresponding natural language sentence is given below.

\begin{itemize}
\item If the ball is in our midfield, position player 3 at (-5, -23).
\item {\it ((bpos (midfield our)) (do (player our {3}) (pos (pt -5 -23))))}
\end{itemize}

\section{Learning Approach}

We adopt the learning model given by \cite{Collins:2005,Collins:2007,Collins:2009} and use it to assign weights to the semantic representations of words. Since a word can have multiple possible syntactic and semantic representations assigned to it, such as $John$ may be represented as $John$ as well as $\lambda x. x @ John$, we use the probabilistic model to assign weights to these representations.

The main differences between our algorithm and the one given by \cite{Collins:2005} are the way in which new semantic representations are obtained. While \cite{Collins:2005} uses a predefined table to obtain these, we obtain the new semantic representations by using inverse $\lambda$ operators and generalization.

\subsection{Learning model and parsing}

We assume that complete syntactic parses are available\footnote{A sentence can have several different parses.}. The parsing uses a probabilistic combinatorial categorial grammar framework similar to the one given by \cite{Collins:2005}. We assume a probabilistic categorial grammar (PCCG) based on a log linear model. Let $S$ denote a sentence, $L$ denote the semantic representation of the sentence, and $T$ denote it's parse tree. We assume a mapping $\bar{f}$ of a triple $(L,T,S)$ to feature vectors $R^d$ and a vector of parameters $\bar{\Theta} \in R^d$ representing the weights. Then the probability of a particular syntactic and semantic parse is given as:

\begin{center}
$P(L,T|S;\bar{\Theta}) = \frac{e^{\bar{f}(L,T,S).\bar{\Theta}}}{\sum_{(L,T)} e^{\bar{f}(L,T,S).\bar{\Theta}}}$
\end{center}

We use only lexical features. Each feature $f_j$ counts the number of times that the lexical entry is
used in $T$.

Parsing a sentence under PCCG includes finding $L$ such that $P(L|S;\bar{\Theta})$ is maximized.

\begin{center}
$argmax_L P(L|S;\bar{\Theta}) = $

$argmax_L \sum_T{P(L,T|S;\bar{\Theta})}$
\end{center}

We use dynamic programming techniques to calculate the most probable parse for a sentence.

\subsection{The inverse $\lambda$ operators}

For lack of space, we present only one of the two Inverse $\lambda$ operators, $Inverse_L$ and $Inverse_R$ of \cite{Marcos:thesis}. The objective of these two algorithms is that given typed $\lambda$-calculus formulas H and G, we want to compute the formula $F$ such that $F@G=H$ and $G@F=H$. First, we introduce the different symbols used in the algorithm and their meaning :

\begin{itemize}
\item Let $G$, $H$ represent typed $\lambda$-calculus formulas, $J^1$,$J^2$,...,$J^n$ represent typed terms, $v_1$ to $v_n$, $v$ and $w$ represent variables and $\sigma_1$,...,$\sigma_n$ represent typed atomic terms.
\item Let $f()$ represent a typed atomic formula. Atomic formulas may have a different arity than the one specified and still satisfy the conditions of the algorithm if they contain the necessary typed atomic terms.
\item Typed terms that are sub terms of a typed term J are denoted as $J_i$.
\item If the formulas we are processing within the algorithm do not satisfy any of the $if$ conditions then the algorithm returns $null$.
\end{itemize}

\begin{definition}[operator :]
Consider two lists of typed $\lambda$-elements A and B, $(a_i,...,a_n)$ and $(b_j,...,b_n)$ respectively and a formula $H$. The result of the operation $H(A:B)$  is obtained by replacing $a_i$ by $b_i$, for each appearance of A in H.
\end{definition}

Next, we present the definition of an inverse operators\footnote{This is the operator that was used in this implementation. In a companion work we develop an enhancement of this operator which is proven sound and complete.} $Inverse_R(H,G)$: \\

\begin{definition}[$Inverse_R(H,G)$]
The function $Inverse_R(H,G)$, is defined as:
\newline
\noindent Given $G$ and $H$:
\begin{it}
\begin{enumerate}
\item If $G$ is $\lambda v.v@J$, set $F = Inverse_L(H,J)$
\item If $J$ is a sub term of $H$ and G is $\lambda v.H(J:v)$
then  $F$ = $J$.
\item If G is not $\lambda v.v@J$, $J$ is a sub term of $H$ and G is $\lambda w.H(J(J_1,...,J_m):w@J_p,...,@J_q)$ with 1 $\leq$ p,q,s $\leq$ m.
then $F$ = $\lambda v_1,...,v_s.J(J_1,...,J_m:v_p,...,v_q)$.
\end{enumerate}
\end{it}
\end{definition}

The function $Inverse_L(H,G)$ is defined similarly.\\



\noindent
{\bf Illustration:}
\noindent\textbf{$\mathbf{Inverse_R}$ - Case 3:}\newline
Suppose $H$ = $in(river,Texas)$ and $G$ = $\lambda v. v@Texas@river$\newline
G is not of the form $\lambda v.v@J$ since $J$ = $Texas@river$ is not a formula. Thus the first condition is not satisfied. Similarly, there is no $J$ that satisfies the second condition. Thus let us try to find a suitable $J$ that satisfies third condition. If we take $J_1$ = $river$ and $J_2$ = $Texas$, then the third condition is satisfied by $G$ = $\lambda x.H((J(J_1,J_2):x@J_2@J_1)$, which in this case corresponds to $G$ = $\lambda  x.H(in(river,Texas):x@Texas@river)$.
Thus, $F$ = $\lambda v_1,v_2.J(J_1,J_2:v_2,v_1)$ and so $F$ = $\lambda v_1,v_2.in(v_2,v_1)$. \newline It is easy to see that $G$ @ $F$ = $H$.

\subsection{Generalization}

Using $INVERSE\_L$ and $INVERSE\_R$, we are able to obtain new semantic representations of particular words in the sentence. However, without any form of generalization, we are not able to extend these to words beyond the ones actually contained in the training data. Since our goal is to go beyond that, we strive to generalize the new semantic representations beyond those words.

To extend our coverage, a function that will take any new learned semantic expressions and the current lexicon and will try to use them to obtain new semantic expressions for words of the same category has to be designed. It will use the following idea. Consider the non-transitive verb ``fly'' of category $S \backslash NP$. Lets assume we obtain a new semantic expression for ``fly'' as $\lambda x. fly(x)$ using $INVERSE\_L$ and $INVERSE\_R$. The $GENERALIZE$ function looks up all the words of the same syntactic category, $S \backslash NP$. It then identifies the part of the semantic expression in which ``fly'' is involved. In our particular case, it's the subexpression $fly$. It then proceeds to search the dictionary for all the words of category $S \backslash NP$. For each such word $w$, it will add a new semantic expression $\lambda x. w(x)$ to the dictionary. For example for the verb ``swim'', it would add $\lambda x. swim(x)$.


%
%

However, the above idea also comes with a drawback. It can produce a vast amount of new semantics representations that are not necessary for most of the sentences, and thus have a negative impact on performance. Thus instead of applying the above idea on the whole dictionary, we perform generalization ``on demand''. That is, if a sentence contains words with unknown semantics, we look for words of the same category and use the same idea to find their semantics. Let us assume $IDENTIFY(word, semantics)$ identifies the parts of $semantics$ in which $word$ is involved and $REPLACE(s, a, b)$ replaces $a$ with $b$ in $s$. We assume that each lexical entry is a triple $(w,cat,sem)$ where $w$ is the actual word, $cat$ is the syntactic category and $sem$ is the semantic expression corresponding to $w$ and $cat$.

{\it $GENERALIZE_D(L,\alpha)$ }

\begin{itemize}
\item For each $l_j \in L$
\begin{itemize}
\item If  $l_j(cat) = \alpha(cat)$
\begin{itemize}
\item $I = IDENTIFY(l_j(w), l_j(sem))$
\item $S = REPLACE(l_j(sem), I, \alpha(w))$
\item $L = L \cup (\alpha(w), \alpha(cat), S)$
\end{itemize}
\end{itemize}

\end{itemize}

As an example, consider the sentence ``Give me the largest state.'' from Table \ref{tab:ex-John2}. Let us assume that the semantics of the word ``largest'' as well as ``the'' is not known, however the semantics of ``longest'' is given by the dictionary as $\lambda x. \lambda y. longest(y, x@y)$. Normally, the system would be unable to parse this sentence and would continue on. However, upon calling $GENERALIZE_D(L,$``$largest$''$)$, the word longest is found in the dictionary with the same syntactic category. Thus this function takes the semantic representation of ``longest'' $\lambda x. \lambda y. longest(y, x@y)$, modifies it accordingly for largest, giving $\lambda x. \lambda y. largest(y, x@y)$ and stores it in the lexicon. After that, the $INVERSE_L$ and $INVERSE_R$ can be applied to obtain the semantics of ``the''.

\subsection{Trivial inverse solutions}

Even with on demand generalization, we might still be missing large amounts of semantics information to be able to use $INVERSE_L$ and $INVERSE_R$. To make up for this, we allow trivial solutions under certain conditions. A trivial solution is a solution, where one of the formulas is assigned a $\lambda x. x$ representation. For example, given $H$, we are looking for $F$ such that $H = G @ F$. If we set $G$ to be $\lambda x. x$, then trivially $F = H$. Thus we can try to carefully set some unknown semantics of words as $\lambda x. x$ which will allow us to compute the semantics of the remaining words using $INVERSE_L$ and $INVERSE_R$. The question then becomes, when do we allow these? In our approach, we allow these for words that do not seem to have any contribution to the final semantic meaning of the text. In some cases, articles such as ``the'', while having a specific place in the English language, might not contribute anything to the actual meaning representation of the sentence. In general, any word not present in the final semantics is a potential candidate to be assigned the trivial semantic representation $\lambda x. x$. These are added with very low weights compared to the semantics found using $INVERSE_L$ and $INVERSE_R$, so that if at one point a non-trivial semantic representation is found, the system will attempt to use it over the trivial one.

As an example, consider again the sentence ``Give me the largest state.'' from Table \ref{tab:ex-John2} with the semantics $answer(A,largest(A, state(A)))$. Let us assume the semantic representations of ``the'' and ``largest'' are not known. Under normal circumstances the algorithm would be unable to find the semantics of ``largest'' using $INVERSE_L$ and $INVERSE_R$ as it is missing the semantics of ``the''. However, as ``the'' is not present in the desired semantics, the system will attempt to assign $\lambda x. x$ as its semantic representation. After doing that, $INVERSE_L$ and $INVERSE_R$ can be used to compute the semantic representation of ``largest'' as $\lambda x. \lambda y. largest(y, x@y)$.

\subsection{The overall learning algorithm.}

The complete learning algorithm used within our approach is shown below. The input to the algorithm is an initial lexicon $L_0$ and a set of pairs $(S_i,L_i), i=1,...,n$, where $S_i$ is a sentence and $L_i$ its corresponding logical form. The output of the algorithm is a PCCG defined by the lexicon $L_T$ and a parameter vector $\Theta_T$.

The parameter vector $\Theta_i$ is updated at each iteration of the algorithm. It stores a real number for each item in the dictionary. The initial values were set to $0.1$.  The algorithm is divided into two major steps, lexical generation and parameters update. The goal of the algorithm is to extract as much information as possible given the provided training data.

In the first step, the algorithm iterates over all the sentences $n$ times and for each sentence constructs a syntactic and (potentially incomplete) semantic parse tree. Using the semantic parse tree, it then attempts to obtain new $\lambda$-calculus formulas by traversing the tree and performing regular applications and inverse computations where possible. Any new semantics are then generalized and stored in the lexicon.

The main reason to iterate over all the sentences $n$ times is to extract all the possible information given the current parameter vector. There may be cases where the information learned from the last sentence can be used to learn additional information from the third sentence, which can then be used to learn new semantics from the second sentence etc. By looping over all sentences $n$ times, we ensure we capture and learn as much information as possible.

Note that the semantic parse trees of the sentences may change once the parameters of words change. Thus even though we are looping over all the sentences $T$ times, the semantic parse tree of a sentence might change as a result of a change in the parameter vector. This change can be very minor, such as change in the semantics of a single word, or in a rare case a major one where most of the semantic expressions present in the tree change. Thus we might learn different semantics of words given different parameter vectors.

In the second step, the parameter vector $\Theta_i$ is updated using stochastic gradient descent. Steps one and two are performed $T$ times. In our experiments, the value of $T$ ranged from $50$ to $100$.

Overall, steps one and two form an exhaustive search which optimizes the log-likelihood of the training model.

\begin{itemize}
\item {\bf Input:}

A set of training sentences with their corresponding desired representations $S = \{(S_i,L_i) : i = 1 . . . n\}$ where
$S_i$ are sentences and $L_i$ are desired expressions. Weights are given an initial value of $0.1$.

An initial lexicon $L_0$. 
An initial feature vector $\Theta_0$.

\item {\bf Output:}

An updated lexicon $L_{T+1}$. 
An updated feature vector $\Theta_{T+1}$.

\item {\bf Algorithm:}

\begin{itemize}

\item For t = 1 . . . T

\item Step 1: (Lexical generation)

\item For i = 1...n.
\begin{itemize}
\item For j = 1...n.

\item Parse sentence $S_j$ to obtain $T_j$
\item Traverse $T_j$
\begin{itemize}
\item  apply $INVERSE\_L$, $INVERSE\_R$ and $GENERALIZE_D$ to find new $\lambda$-calculus expressions of words and phrases $\alpha$.
\end{itemize}

\item Set $L_{t+1} = L_t \cup \alpha$

\end{itemize}
\item Step 2: (Parameter Estimation)

\item Set $\Theta_{t+1} = UPDATE(\Theta_t, L_{t+1})$\footnote{For details on $\Theta$ computation, please see the work by \cite{Collins:2005}}
\end{itemize}
\item return $GENERALIZE(L_T, L_T), \Theta(T)$
\end{itemize}

\section{Experimental Evaluation}

\subsection{The data}

To evaluate our algorithm, we used the standard corpus in GEOQUERY and CLANG. The GEOQUERY corpus contained 880 English sentences with respective database queries. The CLANG corpus contained 300 entries specifying rules, conditions and definitions in CLANG. The GEOQUERY corpus contained relatively short sentences with the sentences ranging from four to seventeen words of quite similar syntactic structure. The sentences in CLANG are much longer, with more complex structure with length ranging from five to thirty eight words.

For our experiments, we used the $C\&C$ parser of \cite{CCG} to provide syntactic parses for sentences. For CLANG corpus, the position vectors and compound nouns with numbers were pre-processed and consequently treated as single noun.

Our experiments were done using a 10 fold cross validation and were conducted as follows. A set of training and testing examples was generated from the respective corpus. These were parsed by the $C\&C$ parser to obtain the syntactic tree structure. These together with the training sets containing the training sentences with their corresponding semantic representations (SRs) and an initial dictionary was used to train a new dictionary with corresponding parameters. This dictionary was generalized with respect of all the words in the test sentences. Note that it is possible that many of the words were still missing their SRs. This dictionary was then used to parse the test sentences and highest scoring parse was used to determine precision and recall. Since many words might have been missing their SRs, the system might not have returned a proper complete semantic parse.

To measure precision and recall, we adopted the measures given by \cite{Mooney:2009}. {\it Precision} denotes the percentage of of returned SRs that were correct, while {\it Recall} denotes the percentage of test examples with pre-specified SRs returned. {\it F-measure} is the standard harmonic mean of precision and recall. For database querying, an SR was considered correct if it retrieved the same answer as the standard query. For CLANG, an SR was correct if it was an exact match of the desired SR, except for argument ordering of conjunctions and other commutative predicates. Additionally, a set of additional experiments was run with ``(definec'' and ``(definer'' treated as being equal.

We evaluated two different version of our system. The first one, $INVERSE$, uses $INVERSE_L$ and $INVERSE_R$ and regular generalization which is applied after each step. The second version, $INVERSE+$, uses trivial inverse solutions as well as on demand generalization. Both systems were evaluated on the same data sets using 10 fold cross validation and the $C\&C$ parser using an equal number of train and test sentences, randomly chosen from their respective corpus. The initial dictionary contained a few nouns, with the addition of one randomly selected word from the set $\{what, where, which\}$ in case of GEOQUERY. For CLANG, the initial dictionary also contained a few nouns, together with the addition of one randomly selected word from the set $\{if, when, during\}$. The learning parameters were set to the values used by \cite{Collins:2005}.

\subsection{Results}

We compared our systems with the performance results of several alternative systems for which the performance data is available in the literature. In particular, we used the performance data given by \cite{Mooney:2009}. The systems that we compared with are: The SYN0, SYN20 and GOLDSYN systems by \cite{Mooney:2009}, the system SCISSOR by \cite{Mooney:2005}, an SVM based system KRIPS by \cite{Mooney:2006}, a synchronous grammar based system WASP by \cite{Mooney:2007}, the CCG based system by \cite{Collins:2007} and the work by \cite{Lu:2008}. Please note that many of these approaches require different parsers, human supervision or other additional tools, while our approach requires a syntactic parse of the sentences and an initial dictionary.

Our and their reported results for the respective corpora are given in the Tables \ref{table-resGeo} and \ref{table-resClang}.

\begin{minipage}[b]{.40\textwidth}
\centering
\begin{small}
\begin{tabular}{|c| c| c| c|}
\hline
 & Precision & Recall & F-measure \\
\hline
INVERSE+  &  93.41 & 89.04 & 91.17\\
INVERSE   &  91.12 & 85.78 & 88.37\\
\hline
GOLDSYN   &  91.94 & 88.18 & 90.02\\
\hline
WASP      &  91.95 & 86.59 & 89.19\\
\hline
Z\&C      &  91.63 & 86.07 & 88.76\\
\hline
SCISSOR   &  95.50 & 77.20 & 85.38\\
KRISP     &  93.34 & 71.70 & 81.10\\
Lu at al. &  89.30 & 81.50 & 85.20\\
\hline
\end{tabular}
\end{small}
\captionof{table}{Performance on GEOQUERY.}
\label{table-resGeo}
\end{minipage}\qquad
\begin{minipage}[b]{.40\textwidth}
\centering
\begin{small}
\begin{tabular}{|c| c| c| c|}
\hline
 & Precision & Recall & F-measure \\
\hline
INVERSE+(i)&  87.67 & 79.08 & 83.15\\
INVERSE+   &  85.74 & 76.63 & 80.92\\
\hline
GOLDSYN   &  84.73 & 74.00 & 79.00\\
SYN20     &  85.37 & 70.00 & 76.92\\
SYN0      &  87.01 & 67.00 & 75.71\\
\hline
WASP      &  88.85 & 61.93 & 72.99\\
KRISP     &  85.20 & 61.85 & 71.67\\
\hline
SCISSOR   &  89.50 & 73.70 & 80.80\\
Lu at al. &  82.50 & 67.70 & 74.40\\
\hline
\end{tabular}
\end{small}
\captionof{table}{Performance on CLANG.}
\label{table-resClang}
\end{minipage}\qquad


%

The $INVERSE+(i)$ denotes training where ``(definec'' and ``(definer'' at the start of SRs were treated as being equal. The main reason for this was that there seems to be no way to distinguish in between them. Even as a human, we found it hard to be able to distinguish between them.

\subsection{Analysis}

Our testing showed that our method is capable of outperforming all of the existing parsers in F-measure. However, there are parsers which can produce greater precision, such as WASP and SCISSOR on CLANG corpus, however they do at the cost in recall. As discussed by \cite{Mooney:2009}, the GEOQUERY results for SCISSOR, KRISP and Lu's work use a different, less accurate representation language FUNSQL which may skew the results. Also, SCISSOR outperforms our system on GEOQUERY corpus in terms of precision, but at the cost of additional human supervision.

Our system is particularly accurate for shorter sentences, or a corpus where many sentences have similar general structure, such as GEOQUERY. However, it is also capable of handling longer sentences, in particular if they in fact consists of several shorter sentences, such as for example ``If the ball is in our midfield, position player 3 at (-5,-23).'', which can be looked at as ``IF A, B'' where ``A'' and ``B'' are smaller complete sentences themselves. The system is capable of learning the semantics of several basic categories such as verbs, after which most of the training sentences are easily parsed and missing semantics is learned quickly. The inability to parse other sentences mostly comes from two sources. First one is if the test sentence contains a syntactic category not seen in the training data. Our generalization model is not capable of generalizing these and thus fails to produce a semantic parse. The second problem comes from ambiguity of SRs. During training, many words will be assigned several SRs based on the training data. The parses are then ranked and in several cases, the correct SR might not be on the top. Re-ranking might help alleviate the second issue.

Unlike the other systems, we do not make use of a grammar for the semantics of the sentence. The reason it is not required is that the actual semantics is analyzed in computing the inverse lambdas, and the richness of $\lambda$-calculus allows us to compute relatively complex formulas to represent the semantic of words.

We also run examples with increased size of training data. These produced larger dictionaries and in general did not significantly affect the results. The main reason is that as discussed before, once the most common categories of words have their semantics assigned, most of the sentences can be properly parsed. Increasing the amount of training data increases the coverage in terms of the rare syntactic categories, but these are also rarely present in the testing data. The used training sample was in all cases sufficient to learn almost all of the categories. This might not be the case in general, for example if we had a corpus with all of the sentences of a particular length and structure, our method might not be capable of learning any new semantics. In such cases, additional words would have to be added to the initial dictionary, or additional sentences of varying lengths would have to be added.

The $C\&C$ parser of \cite{CCG} was primarily trained on news paper text and thus did have some problems with these different domains and in some cases resulted in complex semantic representations of words. This could be improved by using a different parser, or by simply adjusting some of the parse trees. In addition, our system can be gradually improved by increasing the size of initial dictionary.



\section{Conclusions and Discussion}

We presented a new approach to map natural language sentences to their semantic representations. We used an existing syntactic parser, a novel inverse $\lambda$ operator and several generalization techniques to learn the semantic representations of words. Our method is largely independent of the target representation language and directly computes the semantic representations based on the syntactic structure of the syntactic parse tree and known semantic representations. We used statistical learning methods to assign weights to different semantic representation of words and sentences.

Our results indicate that our approach outperforms many of the existing systems on the standard corpora of database querying and robot command and control.

We envision several directions of future work. One direction is to experiment our system with corpora where the natural language semantics is given
through other Knowledge Representation languages such as answer set programming (ASP)\footnote{A preliminary evaluation with respect to a corpus with newspaper text translated into ASP resulted in a precision of 77\%, recall of 82\% with F-measure at 80 using a much smaller training set.} and temporal logic. We are currently building such corpora. Another direction is to improve the statistical learning part of the system. An initial experimentation with a different learning algorithm shows significant decrease in training time with slight reduction in performance. Finally, since our system uses an initial dictionary, which we tried to minimize by only having a few nouns and one of the query words, exploring how to reduce it further and possibly completely eliminating it is a future direction of research.


\bibliographystyle{chicago}
\bibliography{bib-all}

\end{document}